# Survival Analysis on Structured Data using Deep Reinforcement Learning

Renith G*

HCL Technologies, India

Harikrishna Warrier

HCL Technologies, India

Yogesh Gupta

HCL Technologies, India

## Abstract

Survival analysis is playing a major role in manufacturing sector by analyzing occurrence of any unwanted event based on the input data. Predictive maintenance, which is a part of survival analysis, helps to find any device failure based on the current incoming data from different sensor or any equipment. Deep learning techniques were used to automate the predictive maintenance problem to some extent, but they are not very helpful in predicting the device failure for the input data which the algorithm had not learned. Since neural network predicts the output based on previous learned input features, it cannot perform well when there is more variation in input features. Performance of the model is degraded with the occurrence of changes in input data and finally the algorithm fails in predicting the device failure. This problem can be solved by our proposed method where the algorithm can predict the device failure more precisely than the existing deep learning algorithms. The proposed solution involves implementation of Deep Reinforcement Learning algorithm called Double Deep Q Network (DDQN) for classifying the device failure based on the input features. The algorithm is capable of learning different variation of the input feature and is robust in predicting whether the device will fail or not based on the input data. The proposed DDQN model is trained with limited or lesser amount of input data. The trained model predicted larger amount of test data efficiently and performed well compared to other deep learning and machine learning models.

## I. Introduction

Artificial Intelligence is playing a key role in automating most of the applications and making human life easier. Industries such as Manufacturing sector are serving their end consumer efficiently with the help of this technology. Monitoring the operational life period of a device or an equipment is the core requirement in this sector and has implications for managing the maintenance costs efficiently.

### A. Survival Analysis

Survival Analysis is a process built by statistical procedures which is used to predict whether an event occurs or not based on the given input data or feature. This procedure is mainly followed in industries to prevent any failure of machines or device equipment by predicting in prior the event which is supposed to happen. To perform this analysis, it requires necessary input data from the device for a robust prediction. For our research, we have created a dataset with some input data for a particular device to predict whether the device operates normally without any problem or not. The input features of the device we collected includes parameters like voltage, rotation, pressure, and vibration. Based on these input features we have a binary target variable denoting Normal or Failure case. The event occurring for a failure case is very limited and rare scenario. Hence the dataset with target variable for normal case and failure case is unbalanced. To automate the prediction of a device failure based on input data parameters we need the balance the target class. Deep learning algorithms works efficiently mostly in balanced dataset than in an unbalanced dataset. Other scenario to be noted is that whether the algorithm can handle predicting a failure case of the device in future, since the occurrence is very rare. If the algorithm fails to predict the fault scenario, then it affects the production in manufacturing sector leading to huge loss. Predicting a faulty scenario is complicated since it is difficult to determine what combination of input data will lead to failure case. To solve this kind of scenario, we propose a robust and reliable algorithm based on deep reinforcement learning.

**B. Deep Reinforcement Learning**

Reinforcement learning is a popular technique where an agent explores and learns a completely new environment. The learning of an unknown environment cannot happen at a single glance. Multiple trial and error are processed to discover the given environment. Understanding of the environment by the agent is performed by taking actions and getting feedback or reward for the action taken from the environment. Based on the reward received, the agent learns whether the action taken in a particular state is good or bad. Maximizing the overall reward is the aim of the agent. This can be achieved only through the agents experience by making different trial and error in the environment.

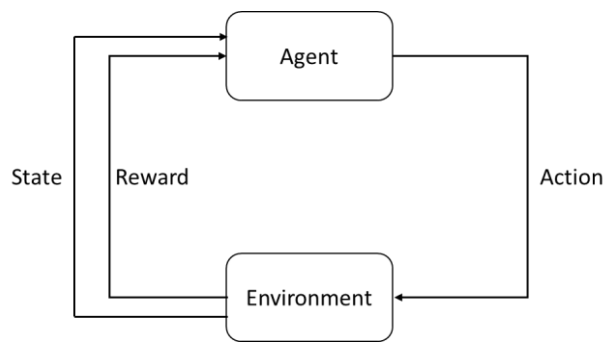

Figure 1: General flow of Reinforcement Learning

The main components of reinforcement learning algorithm are Agent, Environment, State, Action and Reward. Agent is a kind of entity that is used to explore the environment. Environment is a place where agent operates in. The nature of the environment is stochastic since it changes in a random manner. Action is the decision taken by the agent inside the environment. State is that specific part of environment where agent takes the action. Reward is the feedback received from the environment based on the action taken in a state. A good action results in a positive reward, otherwise it leads to a negative reward or penalty. Based on the reward or penalty received for the action taken in a particular state, the agent could understand the unknown environment in a better way. This is the conventional operational process that constitute the reinforcement learning technique as shown in Figure 1. Combination of deep neural network with this traditional reinforcement learning is termed as Deep Reinforcement Learning.

### C. Reinforcement Component with Predictive Dataset

The components of reinforcement learning can be suitable for gaming applications but while implementing them for a tabular data classification problem, the components should be defined in a proper way. The dataset taken for our problem solution has four input features namely voltage, rotation, pressure, and vibration. The resultant output of these input feature provides the target variable as either NORMAL or FAILURE. Now we can relate these given parameters with the components of reinforcement learning. Batch of multiple set of input features constitute the Environment. Each single set of input features represent the State. i.e., single values of voltage, rotation, pressure, and vibration represent State component. The Action represent the target variables NORMAL/FAILURE. Hence two different Action can be taken in a State. Agent is the entity or the algorithm logic to be learned to take corrective action in each state. The Deep Reinforcement Learning algorithm chosen for our research is Deep Q Network (DQN) and Double Deep Q Network (DDQN). The operational flow of these two algorithms varies a little bit, but it holds the property of reinforcement components.

### D. Adaptability for the Input Features not Explored

Reinforcement learning techniques can adapt themselves in various environments. Due to this special feature, they are widely used in gaming applications. Predicting a device failure which occurs rarely is not an easy task for any algorithm due to lack of availability of input data. Deep learning algorithm are the best experts only if sufficient input data is fed into the neural network model. Only with good amount of input data, the algorithm can extract the input features from the given dataset. But if there is a lack of input data for a particular class, then neural network model performs poorly. On the other hand, deep reinforcement learning algorithm can learn in an unknown environment on its own with exploration and exploitation techniques. From the very few input data available to them, reinforcement learning algorithm start extracting the input features by exploring them in a random fashion. This makes the reinforcement learning algorithm to adapt themselves to changes in input features by learning in a continuous fashion.

## II. Related Work

Structured data classification is broadly used in many applications. With the advancement in deep learning techniques, automating many complex analytical tasks has been made easier. Different network architectures have been built and fine-tuned based on the desired problem statement. The ultimate target of the problem statement could vary whether to use classifier to classify the discrete known classes or to use regressor for predicting continuous target variables. Reinforcement learning is more popularly used in gaming application and robot navigation for path planning. Due to random decision learning nature, reinforcement learning is emerging in complex data analysis for text, image, video, audio, and speech as well.

Hankang Sheng et al [1], has proposed a reinforcement learning technique which could perform classification on type of web services present on the internet. The solution can remove the redundant vast data available on the internet and filter the required information for classifying the desired web service. Reinforcement learning can also be able to recognize the important text structure of the language by efficiently implementing the actor critic strategy. This solution is proposed by Feiyang Yang et al [2] for Korean language text by combining the self-attention technique with deep reinforcement learning.

Jiasi Yu et al [3], has implemented solution for discovering the correct textual structure from the text data using reinforcement learning algorithm for classifying the type of the patent. This solution has performed well compared to other baseline deep learning algorithms such as LSTM (Long short-term memory) and BERT transformers. Fengying Yu et al [4], has proposed a solution for improving the quality sentence representation with the help of reinforcement learning method. Natural language processing suffers with basic preprocessing technique called word embeddings. To process the text data efficiently, it is necessary to remove the unwanted information from the text sentence data and extract relevant data from the given text sentence. The author claims that the proposed algorithm compresses the given input text data efficiently by encoding in a proper way using reinforcement method.

Lichen Wang et al [5], has proposed an aspect-based sentiment classification using reinforcement learning algorithm. The popular algorithm used in deep learning for sentiment analysis are the variants of recurrent neural network. But recurrent networks need large amount of labelled data for training otherwise the model fails to generalize and leads to overfitting. The proposed solution by the author is efficient enough to remove the irrelevant words and provides aspect-based sentiment classification which outperforms other deep learning algorithm techniques. Chunyuan Yuan et al [6] proposed a solution for rumor detection in social media contents using reinforcement learning algorithm. Rumor detection in social media is one of the crucial tasks since the content of the topic varies from one another. And building good number of labelled datasets for every rumor content is not quite possible. The author used reinforcement learning for selecting high quality labelled data for performing rumor detection and achieves good result comparatively with other baseline deep learning models.

Qian Li et al [7], has proposed solution for extracting the events and understanding the text sentence based on the arguments. The challenge with implementing deep learning network is that the event varies from one scenario to another. Hence deep learning model strive hard in extracting the events from the text arguments. The author has implemented solution using reinforcement learning by utilizing explicitly the event relationship by exploitation technique used in reinforcement algorithm process. Shubham Jain et al [8], has proposed solution for extracting the important semantic information from unstructured text data by implementing reinforcement learning algorithm. The proposed solution by author proves successful feature extraction by reinforcement method and produce comparatively good result than other deep learning algorithm. Overall reinforcement learning technique seems to be a better option for structured data analysis.

## III. Proposed System

Survival analysis is considered as one of the most critical analytical solution that needs to predict the failure of the equipment more precisely by monitoring the performance based on the input operational data. Automating the prediction of equipment failure need more historical data from the past. This is quite not possible always since input data states which can lead to failure of that equipment is either less frequent or they might be non-existent in an initial dataset. Although there are some solutions developed based on deep learning for unbalanced input datasets, but it has been observed that the models generated in such cases suffers from bias problem due to poor generalization. The solution for this problem can be handled by deep reinforcement learning techniques which is capable of understanding dynamic nature of the input data. Two different deep reinforcement algorithm is considered for our research purpose to solve the given problem statement. i.e., Deep Q Network (DQN) and Double Deep Q Network (DDQN). The resultant output of these algorithms is compared with output of custom build deep learning ANN Network and machine learning LightGBM Network. The proposed deep reinforcement learning implementation solution is proved to be far better than other learning algorithms. Within deep reinforcement algorithm, performance of DDQN seems better than DQN algorithm.

**A. Deep Q Network**

Deep Q Network is an enhancement of traditional Quality Learning (Q-learning) algorithm with combination of deep learning focused to handle higher dimensional complex data. It is important to understand Q-learning algorithm since most of the basic idea emerge from it. Q-learning is modeled in the form of tabular data called as Q-table. Each row in the Q-table is referred as the State and the columns in the Q-table refer to as the Action. The value present in the tabular data is known as Q-values. The Q-values available in each cell of the tabular data represent the kind of action taken for that state. For each action taken in a current state, the Q-value will vary based on the reward received. A good Q-value indicates the chance of getting better rewards in that state-action pair. The traditional Q-learning is useful when the amount of state and action are discrete. The idea behind this algorithm is to find the best optimal policy or solution by approximating the Q-value for the desired state-action pair. The Q-value in the Q-table is updated based on the equation.

It is evident that the Q-learning works for discrete amount of state-action pair. If the amount or volume of state or action is large, then more memory is required for capturing the Q-value for each state-action pair. Another problem arises with higher computation time for exploring the whole environment which is practically not possible. The solution is also not feasible for higher dimensional table where the state and action are continuous and not discrete. To solve this problem, deep reinforcement algorithms such as Deep Q Network (DQN) and Double Deep Q Network (DDQN) is used which combines conventional Q-learning operation with deep learning techniques.

Deep Q Network (DQN) uses neural network to handle higher dimensional state-action pair by acting as a functional approximator to determine the best optimal policy. There is no operational difference between Q-learning and Deep Q Network, both work in the same way. DQN utilizes neural network layers instead of Q-table that is used in Q-learning approach. Based on current state and action as input, the Q-value is determined in Q-learning process. But in Deep Q learning approach, only with current state as the input, all possible Q-values for every action in that current state is determined. The difference of conventional Q-Learning and Deep Q Learning I shown in Figure 2.

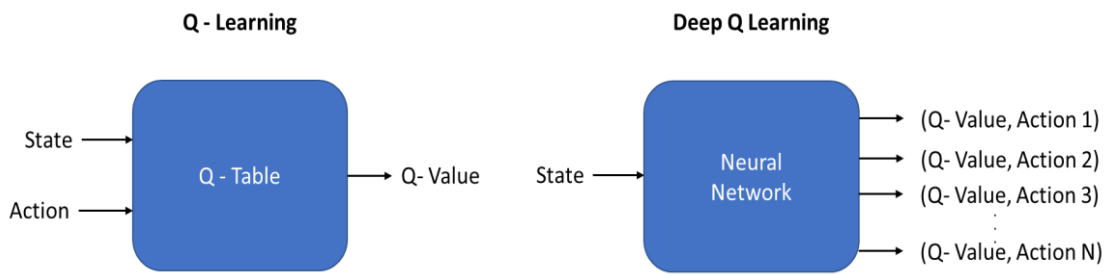

Figure 2: Difference in Input and Output Variables in Q-Learning and Deep Q-Learning

The algorithm or the agent has zero knowledge about the environment. The environment is nothing but the problem to be solved. Agent needs to explore the environment by taking random action in each current state to determine the Q-value of the current state and action taken. After exploring the environment with random actions, the agent should be capable of learning about the environment to take better actions for each state in the future. In this way, the agent learns a policy from its exploration on the environment. After some exploration, the agent needs to start exploiting the agent by enhancing already learned policy. The exploitation of the agent helps to understand the previous learned policy and verify whether optimal solution is found or not. The updating of Q-value always depends on the Q-value of the current state, action and the expected maximum reward from the next possible state Q-value.

$$Q_{Loss} = R_{t+1} + \gamma \max Q(s_{t+1}, a) - Q(s_t, a_t) \qquad (1)$$

The current Q-value can be updated

$$Q(s_t, a_t) = Q(s_t, a_t) + \alpha\, Q_{Loss} \qquad (2)$$

Equation (1) shows the process to compute loss function. $Q(s_t, a_t)$ is the current state action pair Q-value. $\max Q(s_{t+1}, a)$ is the maximum Q-value from the next possible state. $R_{t+1}$ is the maximum expected reward for the action taken. $\gamma$ is the discount factor which plays an important role in maximizing the reward. Equation (2) represents the updating of the Q-value from the calculated loss value. $\alpha$ is the learning rate used for converging the model to achieve best optimal policy or solution. In Deep Q Network, instead of Q-table, the neural network is used. The process flow of DQN algorithm is shown in the Figure 3.

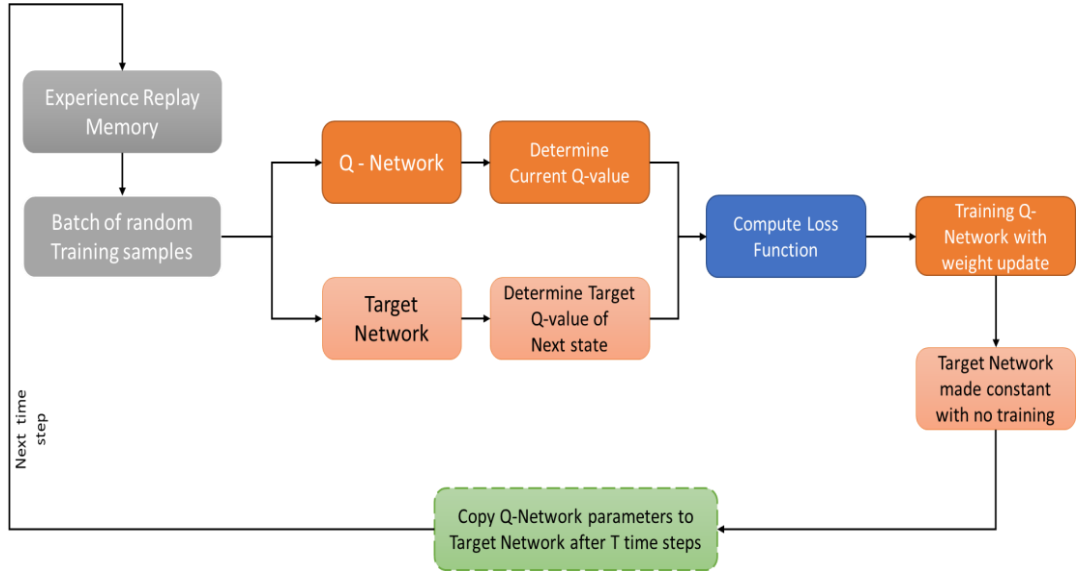

Figure 3: Architecture Flow of DQN

Experience Replay is a memory which is capable of handling the training data to control the huge amount state-action pairs. Experience replay generates or creates training samples by interacting with the environment. It plays random greedy action for the current state and gets the reward and next state. This observation has been stored as single training sample in an array form of quadruple namely – (current state, action, reward, next state). Series of batch of multiple training samples constitute the training data in experience replay memory. Random batch of training samples are taken from the experience replay memory and fed to both Q-Network and Target Network. As each training sample has quadruple values such as (current state, action, reward, next state), the Q-Network takes current state and action as input to predict the current Q-value. Target

Network takes next state value to fetch best Q-value from all possible action from next state as illustrated in the Equation (1). This fetched best Q-value is called as Target Q-value.

With the attained current Q-value, Target Q-value and the Reward, the loss function ($Q_{Loss}$) is computed. Only the Q-Network is trained from the computed loss value and model weights are updated. Target network is not trained and kept constant. This whole process takes place in a single time step. The reason for not training the Target network is to prevent updating of next state value if the weights are updated frequently. Updating weight in each time step will corrupt the actual direction of predicting the target Q-values and makes the network unstable. Although Target Q-values are predicted values and not the actual values, the updated weighted parameters of Q-Network are copied to the Target Network after considerable amount of time steps within an episode.

With the network weight updated for Q-Network in a single time step, again it fetches random training samples from the experience replay memory for training next batch of data. The batch of training samples are shuffled at each time step, so that the algorithm learns diverse range of input data to make the network more robust at each episode. Once the model is trained well, the network can predict with best optimal policy. For final prediction, Q-Network is only used, and Target Network is discarded since it had served its purpose by helping the Q-Network to learn a best optimal policy for the given problem or environment.

## B. Double Deep Q Network (DDQN)

Double Deep Q Network (DDQN) is an enhancement algorithm technique that solves some of the issues faced in DQN algorithm. The architecture flow of DDQN is comparatively similar to the DQN network, except choosing the maximum Q-value from the next state. In DQN network, the estimation of Q-value is based on random choosing of action in a state initially. The Q-values are updated or increased based on the action chosen in that state. After several iteration, the algorithm choses the action depending on the maximum Q-value in that state. The highest Q-value is considered as best action of that state. The chosen Q-value may not be the optimal policy due to poor exploration of agent in the given environment. The best action for a state might be another action with lower Q-value than the current maximum Q-value. This is because the agent has not considered the best action earlier due to poor exploration and have not got a chance to choose that action in a state. Now this leads to overestimation of Q-value for the non-best action and increase in bias and noises. To solve this problem, a simple modification is done in retrieving the maximum Q-value from the next state.

$$a_{qnet} = \max Q_{qnet}(s_{t+1}, a) \quad (3)$$

$$Q_{qnet}(s_t, a_t) = R_{t+1} + \gamma Q_{tnet}(s_{t+1}, a_{qnet}) \quad (4)$$

Equation (3) and (4) clearly explains how the Q-value from next state is computed. $Q_{qnet}$ implies Q-Network and $Q_{tnet}$ implies Target Network. $a_{qnet}$ is the best action of Q-Network which possess maximum Q-value from next state shown in Equation (3). Compute the estimated Q-value in the Target network from the best action $a_{qnet}$ chosen in Equation (3). This selection of Q-value from next state tends to reduce bias and noise problem and improve in selecting best Q-value in future steps.

# IV. Results and Discussion

The aim of our research is to check whether the algorithm developed has potential of predicting a wide range of test data which are not used while model training. We have taken an equipment or device operational life period based on certain parameters like voltage, rotation, pressure, and vibration. Based on these parameters, the algorithm must predict whether the device will fail or not. Capturing of device data for failure case is very limited since its occurrence is not frequent. One of the most common challenge faced in predictive maintenance problem is that the dataset suffers from unbalanced classes. We have taken a three different device data for our research analysis for a fair comparison. First device data (Device-1) has 8717 Normal class and 44 Failed class with total sum of 8761 data. Second device data (Device-2) has 8720 Normal class and 41 Failed class with total sum of 8761 data. Third device data (Device-3) has 8721 Normal class and 40 Failed class with total sum of 8761 data.

For the algorithm performance calculation, F1-Score has been considered due to unbalanced dataset. With the proposed method, deep reinforcement learning algorithm DDQN and DQN can predict large amount of test data efficiently, even when the amount of training data is less. The main hyperparameters used in both algorithms namely DQN and DDQN are similar. Total number of episodes is 150000, batch size is 32, learning rate is 0.0025, updating target network for every 800 steps, gamma value is zero, minimum epsilon decay value from exploration to exploitation is 0.5. The network layers for both DQN and DDQN algorithm used for Q-Network and Target Network are similar. Network layers constitute of three fully connected layer and final classifier layer. First network layer has 128 nodes, second network layer has 64 nodes, third network layer has 32 nodes and final layer has 2 nodes. First three layers has ReLU as activation function. The 2 nodes in final layer describe number of actions to be taken by the algorithm denoting FAILURE or NORMAL. The final layer does not have activation since it needs to output pure Q-values.

Table 1: Comparison results of different algorithms in Device-1

| Test Data Size | F1 – Score | | |
| --- | --- | --- | --- |
| | 20 % | 50 % | 80 % |
| **RL - DDQN** | 0.8571 | 0.7555 | 0.63 |
| **RL - DQN** | 0.8 | 0.7142 | 0.56 |
| **DL – ANN** | 0.7272 | 0.6315 | 0.5 |
| **ML – Light GBM** | 0.5263 | 0.3684 | 0.3389 |

For an effective analysis, performance of deep reinforcement learning algorithm is compared with other algorithms. The comparison is done for four different algorithms namely Double Q Network and Deep Q Network from deep reinforcement learning techniques, Artificial Neural Network (ANN) from deep learning technique and LightGBM from traditional machine learning perspective. For a fair comparison, common hyperparameters of all the algorithms are made similar. The network layers used in deep reinforcement learning (DDQN & DQN) and deep learning (ANN) are similar.

Batch size of 32 is used in all four algorithms. Learning rate of 0.0025 is common for DDQN, DQN and ANN. The comparison table for algorithm performance result for Device-1 is shown in the Table 1.

The usual training and test data size used for typical algorithm development are 80 % and 20 % respectively. To showcase the effectiveness of reinforcement learning algorithm, the analysis is done in different test data size. The result captured in the Table 1 is highlighted based on various test data size for Device-1. The different size of test data used for analysis is 20 %, 50 %, and 80 %. The ratio of training and test data used are 80:20, 50:50 and 20:80. Validation data of 20 % is used among the training data only. The test data computed for final evaluation is not used during the model training process. From the Table 1, it is evident that deep reinforcement learning algorithm performs better than other neural network and machine learning algorithms.

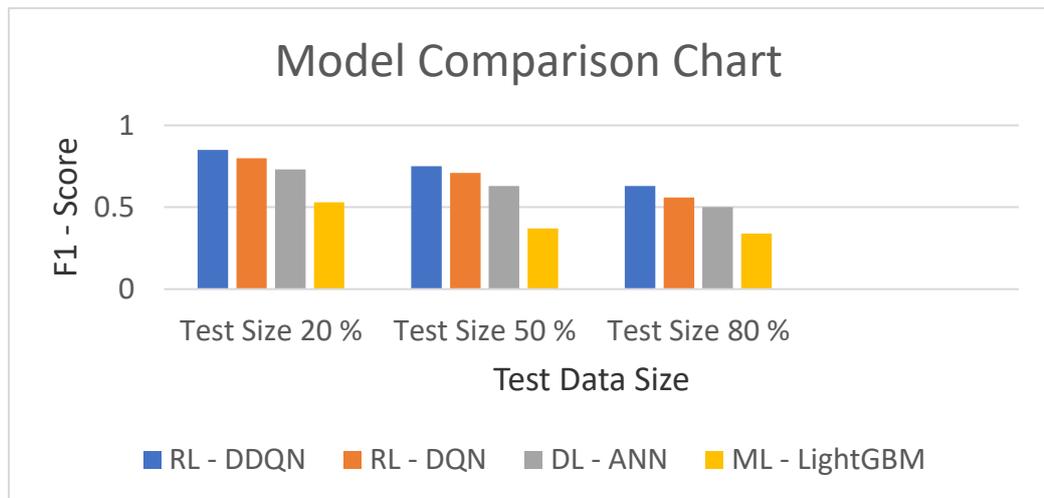

Figure 4: Performance chart showcasing different algorithms in Device-1

Deep reinforcement learning is better even for the prediction of more unseen data with higher test data size. But other algorithm performance such as ANN and LightGBM degrades when the test data size is higher. This can be seen when training data of 20% is trained and evaluated the trained model with 80% of test data. Even with a smaller number of training data, deep reinforcement learning can predict better for unseen data than other algorithms. The comparison chart for performance measures of different algorithms taken for Device-1 can be seen in Figure 4.

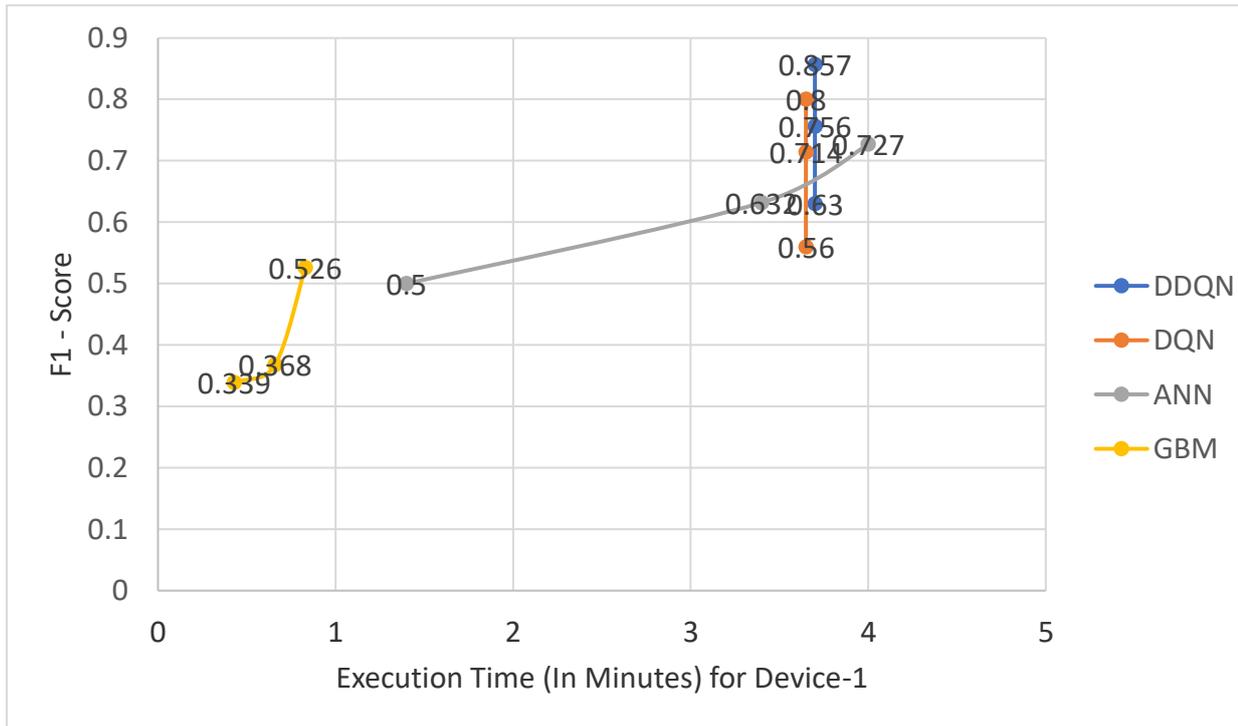

Figure 5: Model training time duration for each algorithm compared with predicted F1-Score in test data

The algorithm implementation and performance analysis done are executed in Intel i5 processor with 16 GB RAM memory. The model training is performed in CPU mode only. The execution time duration comparison with achieved F1 Score metric values for performing model training of each algorithm with different test data size is shown in the Figure 5. Both DDQN and DQN model training execution with good performance result are close and similar for different test data size. ANN algorithm model training time with more training data is close to deep reinforcement learning algorithm, but the performance result of ANN is poor compared to DDQN and DQN. The machine learning LightGBM algorithm execute very quickly but perform poorly in prediction of test data.

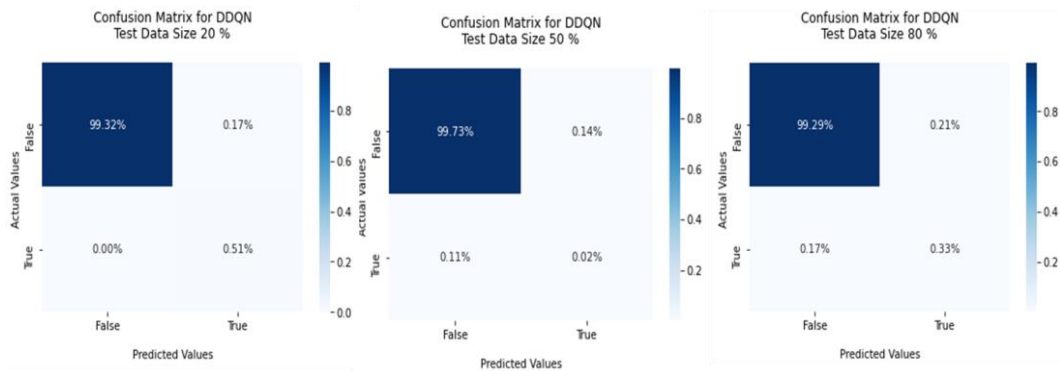

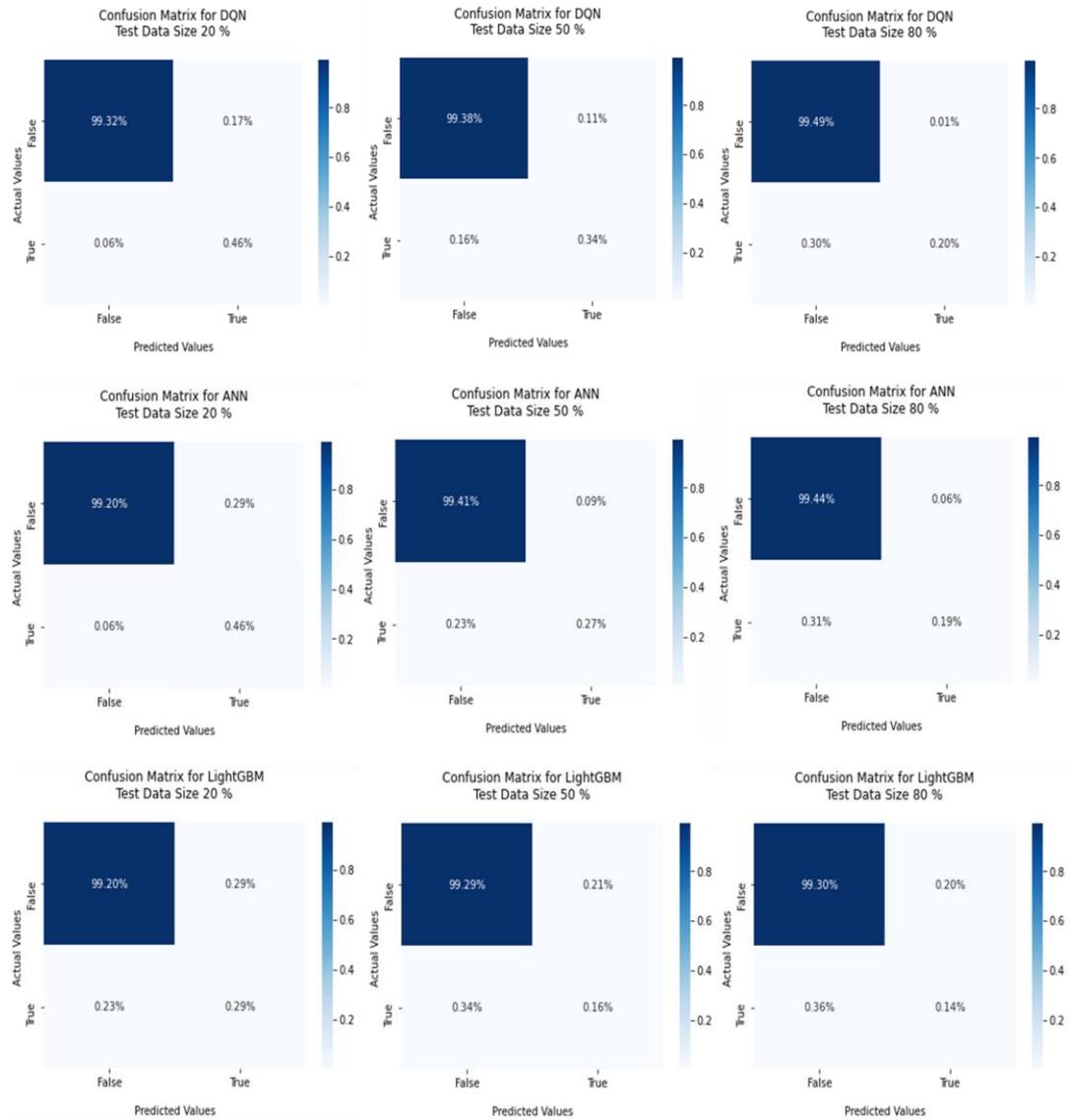

Figure 6: Confusion matrix of algorithms for F1-Score with variant test data size for Device-1

The confusion matrix achieved for algorithms DDQN, DQN, ANN, Light GBM prediction with different test data size of 20%, 50% and 80% is shown in Figure 6. The confusion matrix visually shows how each algorithm performs with various test data which are not seen during model training. Prediction result with test size of 80% implies how reinforcement learning can perform well in varied unseen future data. The range of input feature used in the model training might not be similar with the range of input features used in predicting the unseen test data with the trained model. This variation of input features used in training and test data for Device-1 is shown in Table 2.

| Input Feature | Training Data | | Test Data | |
|---|---|---|---|---|
| | Min Range | Max Range | Min Range | Max Range |
| Volt | 121.989 | 225.1894 | *119.059* | *237.9385* |
| Rotate | 249.9303 | 598.4436 | *215.6194* | *636.3645* |
| Pressure | 69.5447 | 142.8442 | *58.7337* | *155.2575* |
| Vibration | 24.6984 | 67.409 | *22.6668* | 67.6334 |

Table 2: Input Data Feature Range Variation for Device-1

In Table 2, the minimum and maximum range variation of input feature data for test data size of 80% is shown. While looking into the minimum and maximum range of both training and test data, we could see that most of the test data features are beyond the boundary limit of the training data. Minimum range of test data is lesser than minimum range used in training data and vice vera for the maximum range of training and test data. From this result it is evident that the reinforcement learning algorithm classify wide range of variation in input data more precisely than other deep learning and machine learning algorithms.

Similar comparisons are done for second device (Device-2) with same input features namely voltage, rotate, pressure and vibration. The comparison table for algorithm performance result for Device-2 is shown in the Table 3.

Table 3: Comparison results of different algorithms in Device-2

| Test Data Size | F1 – Score | | |
|---|---|---|---|
| | 20 % | 50 % | 80 % |
| **RL - DDQN** | 0.8888 | 0.8636 | 0.6588 |
| **RL - DQN** | 0.8 | 0.7222 | 0.5842 |
| **DL – ANN** | 0.7777 | 0.6111 | 0.5283 |
| **ML – Light GBM** | 0.2857 | 0.1428 | 0 |

Table 3 shows the implementation result in Device-2 for different algorithms namely DDQN, DQN, ANN and Light GBM model architecture with same hyperparameters used for previous analysis. DDQN and DQN algorithm performs better compared to other deep learning and machine learning algorithm. DDQN score good performance result compared to DQN. Overall deep reinforcement learning algorithm can predict wide range of test data efficiently. The execution time taken for each model training along with their test data performance score is shown in Figure 7.

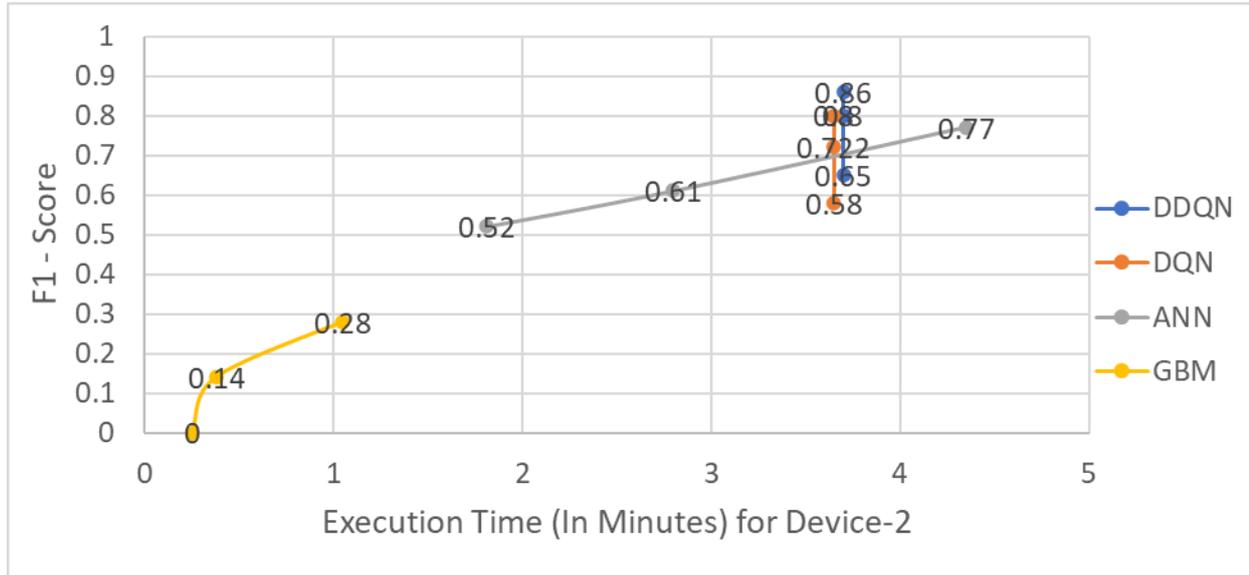

Figure 7: Model training time duration for each algorithm compared with predicted F1-Score in test data

The confusion matrix attained for the Device-2 predicted result in test data for each algorithm with various test data size is shown in Figure 8.

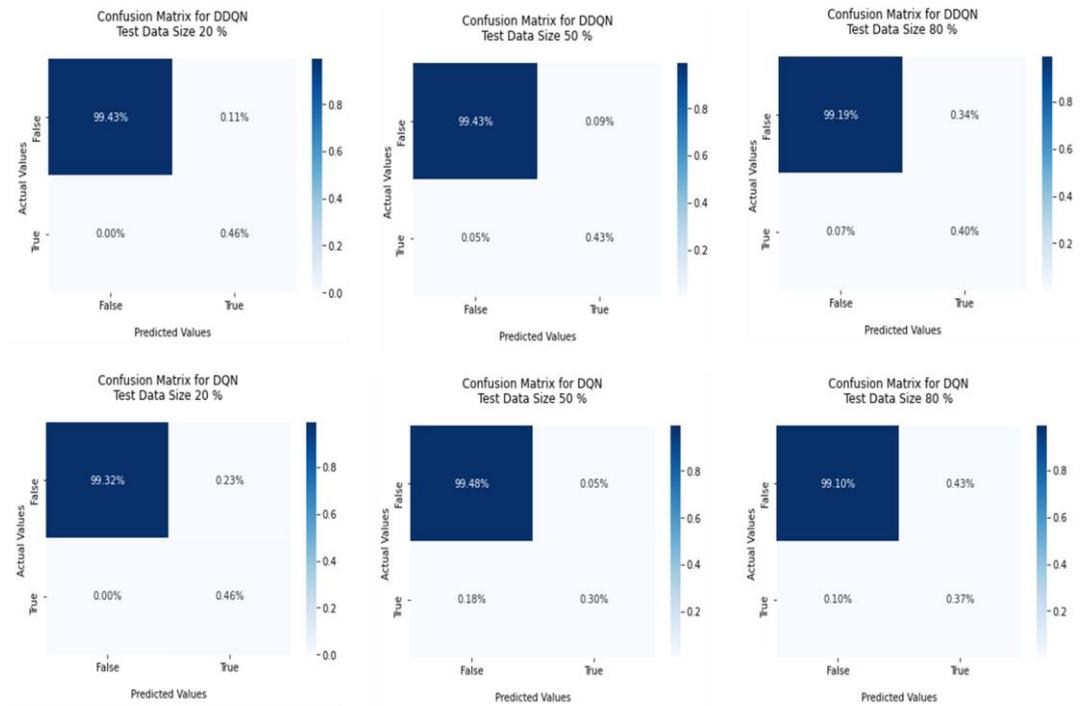

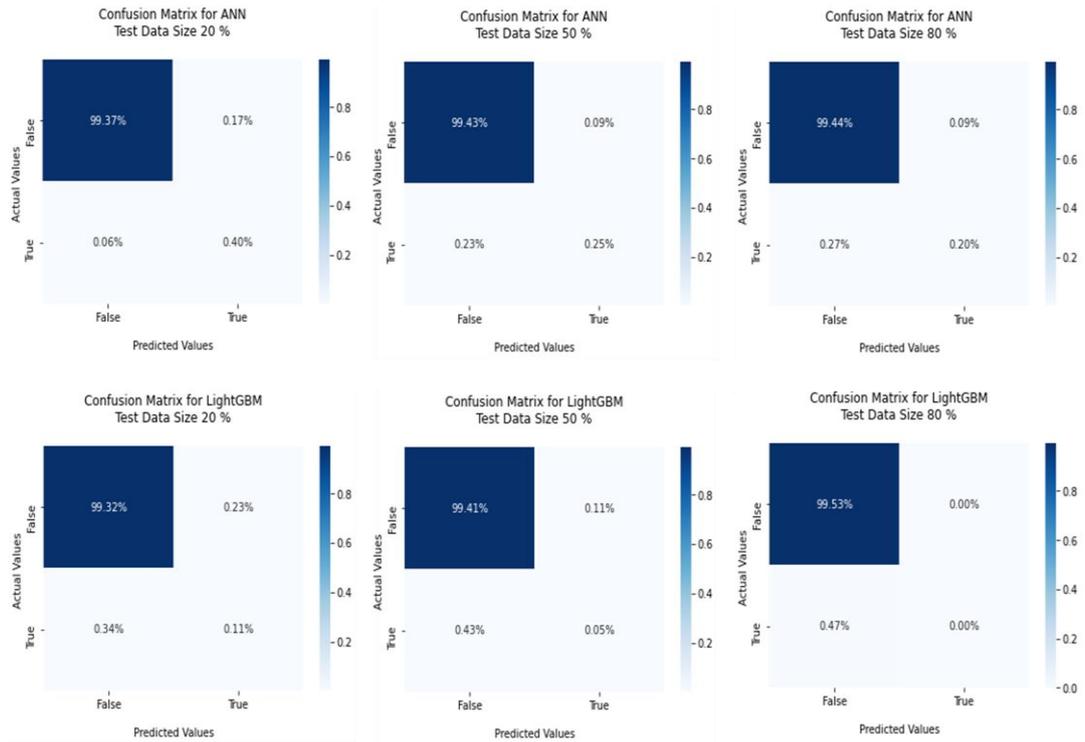

Figure 8: Confusion matrix of algorithms for F1-Score with variant test data size for Device-2

The variation of input data features such as Volt, Rotate, Pressure and Vibration used in both training and test data size for Device-2 is shown in Table 4.

| Input Feature | Training Data | | Test Data | |
| --- | --- | --- | --- | --- |
| | Min Range | Max Range | Min Range | Max Range |
| **Volt** | *100.1941* | 235.4937 | 110.5486 | *241.5164* |
| **Rotate** | 188.362 | 631.5001 | *182.4665* | *638.1777* |
| **Pressure** | 71.3938 | 140.6421 | *63.5371* | *152.0524* |
| **Vibration** | 24.244 | 65.5537 | *19.9554* | *70.3659* |

Table 4: Input Data Feature Range Variation for Device-2

Table 4 explains the minimum and maximum range of input data features used for both training and test data. This range is taken for the model training with test data size of 80% since it has more range variation compared to other training and test data split. From Table 4, it is evident that mot of the minimum and maximum range of test data has more variation compared with training data range. With this amount of test data variation in features, deep reinforcement learning algorithm performs better compared with other deep learning and machine learning algorithms.

Similar comparisons are done for third device (Device-3) with same input features namely voltage, rotate, pressure and vibration. The comparison table for algorithm performance result for Device-2 is shown in the Table 5.

Table 5: Comparison results of different algorithms in Device-3

| Test Data Size | F1 – Score | | |
|---|---|---|---|
| | **20 %** | **50 %** | **80 %** |
| **RL - DDQN** | 0.8571 | 0.7058 | 0.5714 |
| **RL - DQN** | 0.7272 | 0.7647 | 0.4313 |
| **DL – ANN** | 0.5714 | 0.6206 | 0.4489 |
| **ML – Light GBM** | 0.1666 | 0.3333 | 0.1578 |

Table 5 shows the implementation result in Device-3 for different algorithms namely DDQN, DQN, ANN and Light GBM model architecture with same hyperparameters used for previous analysis. DDQN and DQN algorithm performs better compared to other deep learning and machine learning algorithm. DDQN score good performance result compared to DQN for test data size 20% and 80%. DQN performs better than DDQN for test data size 50%. ANN performance score for test data size of 80% is close to DQN algorithm. Overall deep reinforcement learning algorithm can predict wide range of test data efficiently. The execution time taken for each model training along with their test data performance score is shown in Figure 9.

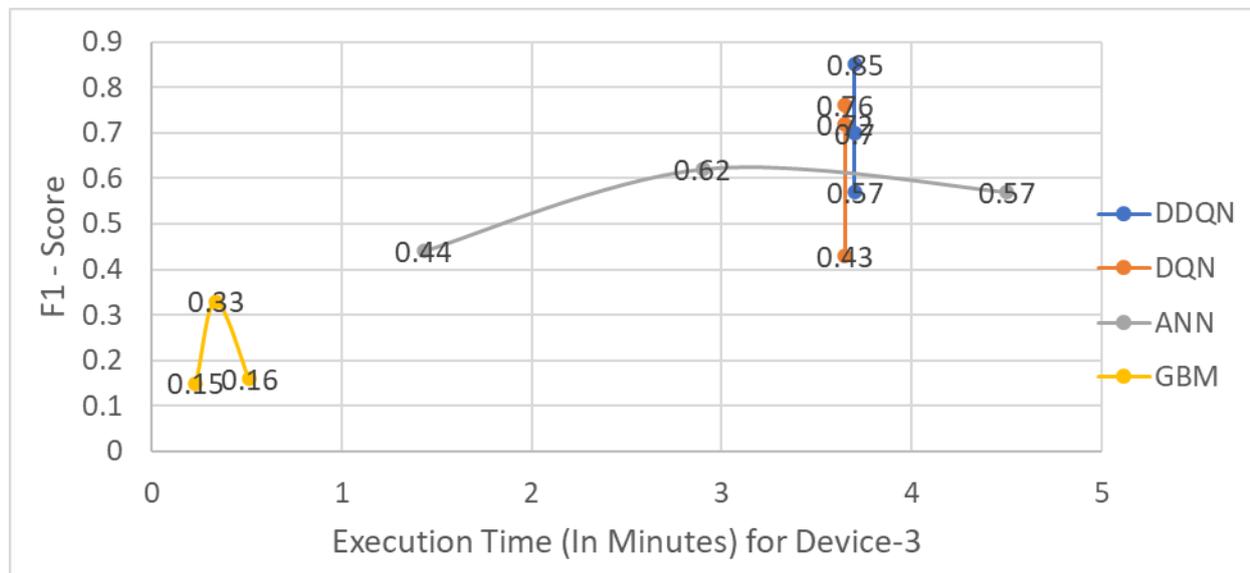

Figure 9: Model training time duration for each algorithm compared with predicted F1-Score in test data

The confusion matrix attained for the Device-3 predicted result in test data for each algorithm with various test data size is shown in Figure 10.

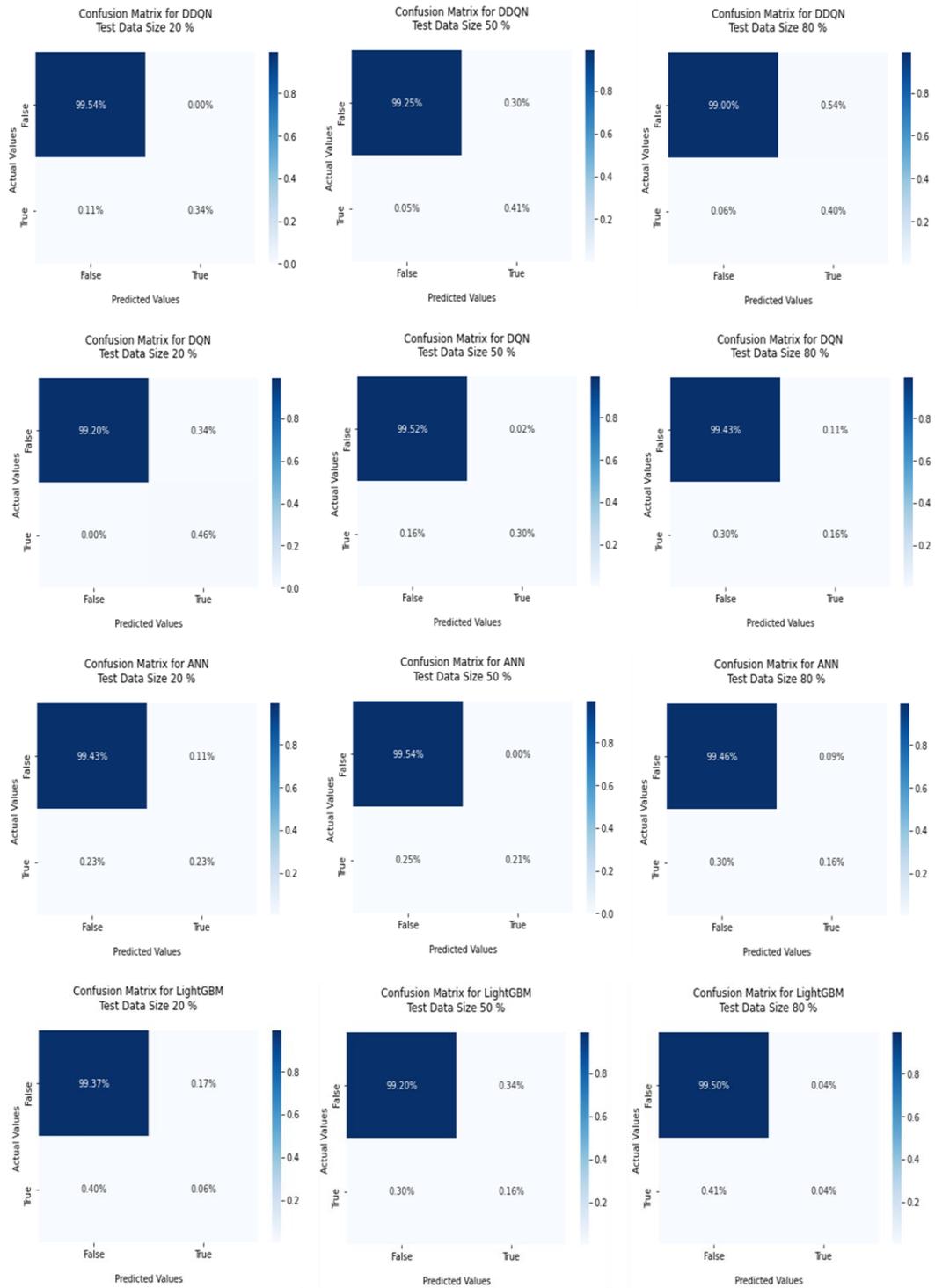

Figure 10: Confusion matrix of algorithms for F1-Score with variant test data size for Device-3

The variation of input data features such as Volt, Rotate, Pressure and Vibration used in both training and test data size for Device-3 is shown in Table 6.

| Input Feature | Training Data | | Test Data | |
|---|---|---|---|---|
| | Min Range | Max Range | Min Range | Max Range |
| Volt | 120.1948 | 224.9753 | *113.8054* | *233.8038* |
| Rotate | 245.3554 | 629.9048 | *236.0646* | *635.1825* |
| Pressure | *61.8601* | 141.1452 | 63.8608 | *149.2234* |
| Vibration | 25.5943 | 63.0628 | *21.2161* | *64.5308* |

Table 6: Input Data Feature Range Variation for Device-3

Table 6 shows the minimum and maximum range of input data feature used in training and test data for Device-3. The input feature include volt, rotate, pressure and vibration. Most of the minimum and maximum range of test data are beyond the boundary of training data except for the minimum range in vibration feature. With this given input feature variation in test data, deep reinforcement algorithm performs better compared with other deep learning and machine learning algorithms.

From the research study and analysis of three different device data implemented in different algorithms such as deep reinforcement learning DDQN, DQN, deep learning algorithm ANN and machine learning algorithm Light GBM, we can conclude that deep reinforcement learning predicting better result for a wide range of variation in unseen test data.

## V. Conclusion and Future Work

Predicting the survival of a device or an equipment based on the input data is a difficult task. This is due to lack of capturing failure data since the occurrence is very rare. Deep Learning algorithms will suffer if there is no sufficient input data available for a particular class label. Neural networks works efficiently only if there is good number of features that can be extracted from the provided input data. The algorithm trained with these unbalanced dataset fails to predict the output, since it does not learn all the correct features of the input class. The solution proposed with deep reinforcement learning algorithm which possess continual learning capability and adapts itself with varied input features, predict the class with lesser data efficiently. With this proposed method, the failure of the equipment is predicted more precisely than other learning algorithms with higher amount of test data which are not used during model training. This also shows that deep reinforcement learning can adapt themselves even for the input feature that are not learned during the training process. As a future work, the same analysis could be implemented in image data modality and the results could be compared in a similar fashion.

# Author Biography

| Photo | Biography |
|---|---|
| 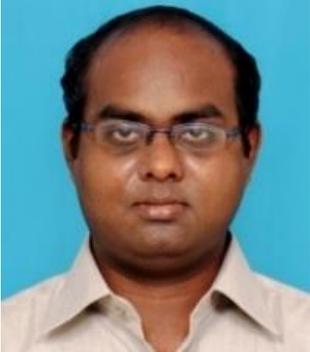 | **Renith G** received his B.E. Degree in Electronics and Communication Engineering from Anna University, India, in 2010 and received M.Tech Degree in Computer science and Engineering from SRM Institute of Science and Technology, India, in 2020. He works with HCL Technologies, India as Technical Lead. His current research interests include Image Processing, Computer Vision and Deep Learning. |
| 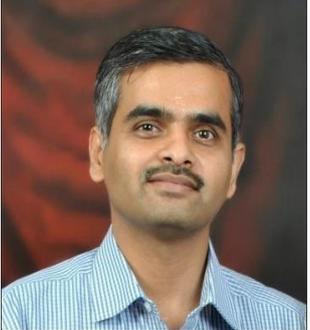 | **Harikrishna Warrier** received his B-Tech in Electronics and Electrical Communications from IIT Kharagpur and has done Post Graduate Certification in Business Management from XLRI Jamshedpur. He works with HCL Technologies, India as Solutions Director. His current research interests include MLOps, Data Centric AI, Edge Analytics and application of AI/ML in 5G. Hari has filed 7 patents in the areas of analytics and wireless communications and has 2 publications in international conferences. |
| 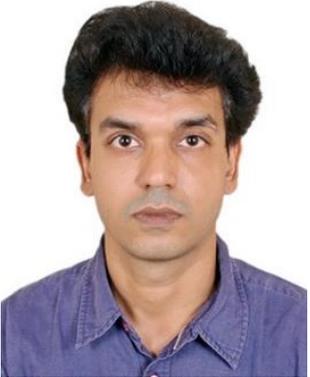 | **Yogesh Gupta** received his Master's degree in Computer Science from IP University, Delhi, India. He works with HCL Technologies, India as a Global Technology Director. His current research interests include MLOps, Federated Learning, Explainable AI and Lake-house architectures. Yogesh has filed 15+ patents in analytics, process automation and cloud areas. |